\documentclass{article}

\usepackage{spconf}
\usepackage{graphicx}
\usepackage{xspace}
\usepackage[acronym]{glossaries}
\usepackage{color, colortbl}
\usepackage{ marvosym }

\newacronym{ai}{AI}{Artificial Intelligence}
\newacronym{bn}{BN}{BatchNorm}
\newacronym{cnn}{CNN}{Convolutional Neural Network}
\newacronym{ch}{CH}{Clever Hans}
\newacronym{deeplift}{DeepLIFT}{Deep Learning Important FeaTures}
\newacronym{dnn}{DNN}{Deep Neural Network}
\newacronym{dtd}{DTD}{Deep Taylor Decomposition}
\newacronym{ebp}{EBP}{Excitation Backpropagation}
\newacronym{gcam}{Grad-CAM}{Grad-Class Activation Mapping}
\newacronym{ggcam}{Guided Grad-CAM}{Guided Grad-Class Activation Mapping}
\newacronym{itg}{I$\times$G}{Input$\times$Gradient}
\newacronym{ig}{IG}{Integrated Gradients}
\newacronym{gbp}{GBP}{Guided Backpropagation}
\newacronym{lrp}{LRP}{Layer-wise Relevance Propagation}
\newacronym{ml}{ML}{machine learning}
\newacronym{nn}{NN}{neural network}
\newacronym{xai}{XAI}{eXplainable Artificial Intelligence}

\newcommand{\ie}{{i.e.}\xspace}

\newcommand{\cf}{{cf.}\xspace}
\newcommand{\wrt}{{w.r.t.}\xspace}
\newcommand{\eg}{{e.g.}\xspace}

\title{Measurably Stronger Explanation Reliability via Model Canonization}

\name{Franz Motzkus %
\qquad Leander Weber %
\qquad Sebastian Lapuschkin%
}
\address{Department of Artificial Intelligence, Fraunhofer Heinrich Hertz Institute, 10587 Berlin, Germany
}

\begin{document}
                                                                                                                                                            
\maketitle

\begin{abstract}
    While rule-based attribution methods have proven  useful for providing local explanations for Deep Neural Networks,
    explaining modern and more varied network architectures
    yields new challenges in generating trustworthy explanations, since the established rule sets might not be sufficient or applicable to novel network structures.
    As an elegant solution to the above issue,
    \emph{network canonization} has recently been introduced.
    This procedure leverages the implementation-dependency of rule-based attributions and restructures a model into a functionally identical equivalent of alternative design to which established attribution rules can be applied.
    However, the idea of canonization and its usefulness have so far only been explored qualitatively.
    In this work, we quantitatively verify the beneficial effects of network canonization to rule-based attributions
    on VGG-16 and ResNet18 models with BatchNorm layers
    and thus extend the current best practices for obtaining reliable neural network explanations.
    \let\thefootnote\relax\footnotetext{This work was supported by
    the European Union's Horizon 2020 research and innovation programme (EU Horizon 2020) as grant [iToBoS (965221)].
    \Letter~\texttt{sebastian.lapuschkin@hhi.fraunhofer.de}
    }
\end{abstract}

\section{Introduction}
In recent years, the field of \gls{xai} has shifted into focus,
with the objective to solve the "black-box"-problem inherent to many (deep) machine learning predictors
in order to better understand the reasoning of these models and consequently increase trust in them.
Current work from the field has highlighted the usefulness of contemporary \gls{xai} approaches for understanding and improving models and datasets in various application scenarios \cite{samek2021explaining}.

Many contemporary explainability methods yield so-called attribution maps,
\ie,~per-input-dimension indicators of how (much) a model,
\eg, a \gls{dnn},
has used the value of a particular input unit during inference \cite{Bach2015Pixel,Shrikumar2017Learning},
or whether the model is sensitive to its change \cite{morch1995visualization,baehrens2010explain}.
This information can be obtained either via perturbation-based approaches treating the model as a black-box~\cite{ribeiro2016should},
based on the gradient~\cite{Simonyan2014Deep,sundararajan2017axiomatic},
or with techniques applying modified backpropagation through the model~\cite{Bach2015Pixel,Shrikumar2017Learning}.
While (modified) backpropagation-based methods are popular choices for large-scale \gls{xai} experiments and practitioners,
\eg, due to their efficiency and expressivity,
rule-based variants such as \glsdesc{lrp} or \glsdesc{ebp} are known to not be implementation invariant \cite{Montavon2019Gradient}:
Rule-based methods assign dedicated backpropagation "rules" to different parts of a neural network,
allowing for individual treatment of specific layers attuned to their function within the model ~\cite{Kohlbrenner2020Towards}.
That is, with \glspl{dnn} representing complex mathematical functions in a structured manner via layers,
a different expression of the very same function
via an alternative or completely new network architecture may lead to different ---
or in the worst case even misrepresentative --- explanation outcomes.
Furthermore, the continuous evolution of neural network architectures requires rule-based methods to regularly adapt to newly introduced network elements such as \gls{bn} \cite{Ioffe2015Batch} layers or skip connections in ResNets \cite{He2016Deep}.

In awareness to these issues, \emph{neural network canonization} has been introduced in previous work \cite{Hui2019Batchnorm,Guillemot2020Breaking}
and constitutes an elegant solution for transforming a variety of highly specialized 
neural network elements into canonical and well-understood, yet functionally equivalent network architectures for improved model explainability,
where established rules are easily applicable.
Thereby, network canonization has the potential for improving a model's architecture (\wrt~its original design) to \emph{fix problems} in an explainability context, and thus \emph{increase the potential} for model trustworthiness without affecting its functionality.
Existing work~\cite{yeom2021pruning} describes canonization as a required step for reliable results, and, as a consequence, the authors of \gls{xai} software even build such model processors into their toolkits~\cite{Anders2021Software}.
Assessments of
network canonization on visualized attribution maps \cite{Hui2019Batchnorm,Guillemot2020Breaking} exist on a qualitative level.
However, to the best of our knowledge, there is no single study dedicated
to the \emph{quantification} of the beneficial effects of canonization on neural network explanations.

In this work, we dedicate efforts to the quantification and numerical verification of the
benefits of canonization on
two popular \gls{dnn} architectures (VGG-16~\cite{Simonyan2015Very} and ResNet18~\cite{He2016Deep} models, in representation of similar architectures)
on data from the ILSVRC2012~\cite{russakovsky2015imagenet} benchmark dataset.
In particular,
with a focus on variants of the popular \glsdesc{lrp}~\cite{Bach2015Pixel} technique
and gradient-based attribution methods,
we canonically extend the work of \cite{Hui2019Batchnorm,Guillemot2020Breaking}
by measurably demonstrating --- at hand of experiments of attribution-guided input perturbation~\cite{Samek2017Evaluating}
and object localization using annotated ground truth bounding boxes~\cite{Kohlbrenner2020Towards} ---
that model canonization is a critical factor for improving the quality and faithfulness of local explanations in rule-based approaches.

\section{Methods}

\subsection{Network Canonization}

\glsfirst{bn} layers \cite{Ioffe2015Batch} have been shown to be effective building blocks for improving and accelerating neural network training.
By eliminating the internal covariate shift in the activations, 
they stabilize the  distribution of intermediate representations and improve the gradient flow through the network, mitigating issues such as gradient shattering or -explosion \cite{Ioffe2015Batch}. 
In some modified backpropagation methods, 
such as \gls{lrp}, 
it is not inherently clear how to adequately propagate attribution scores through \gls{bn} layers. 
An elegant solution for avoiding this issue altogether is to apply model canonization to these layers
\cite{Hui2019Batchnorm, Guillemot2020Breaking, Binder2020Notes}. 
That is, a model (or parts thereof) is restructured in such a way that the result is functionally equivalent in terms of inference behavior for all possible input samples,
but only contains layer types supported by well-understood attribution propagation rules.
For this purpose, 
\gls{bn} layers are commonly resolved by merging their parameters with an adjacent linear (\ie, convolutional or fully-connected) layer \cite{Guillemot2020Breaking, Binder2020Notes} into a single and functionally equivalent linear layer with fused parameters, replacing both original layers.

\begin{figure}[t]
    \centering
    \includegraphics[width=.9\linewidth]{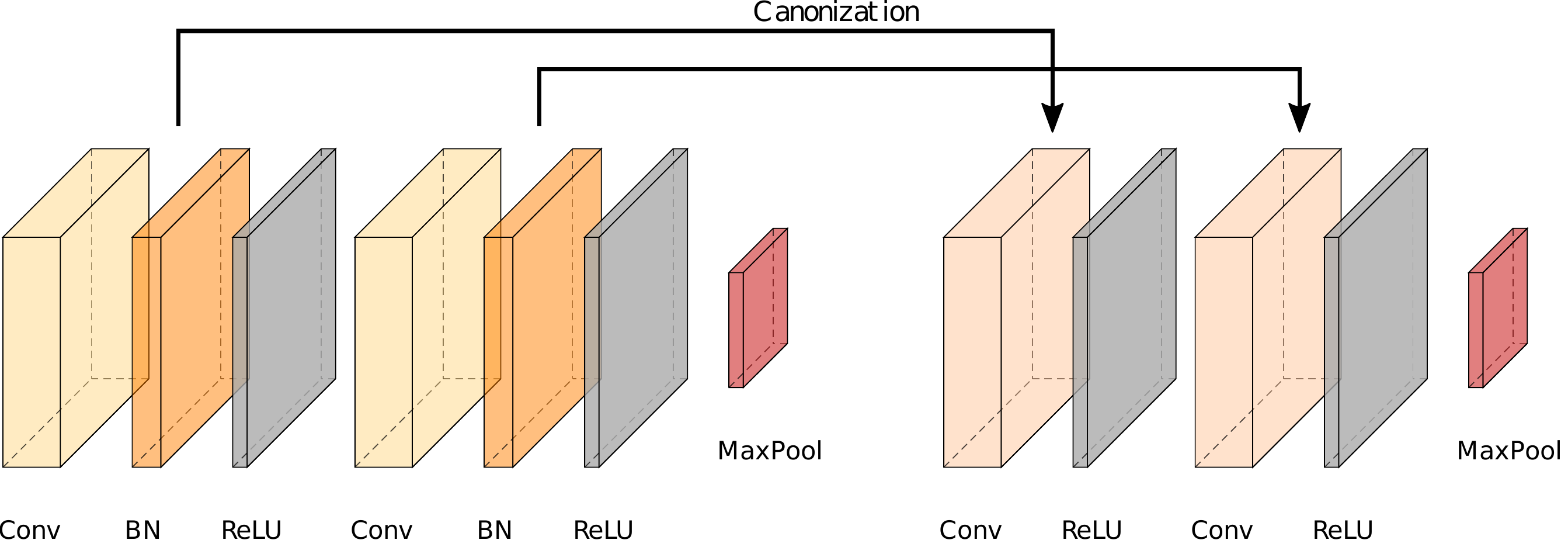}
    \caption{Exemplary canonization of a VGG-16 model segment with \gls{bn} layers.
    As the \gls{bn} is applied before the activation, the parameters of the convolution and the \gls{bn} are fused to build a new convolutional layer. The \gls{bn} layer is then omitted.
    }
    \label{fig:canonization}
\end{figure}
A schematic example for canonization is given in Figure~\ref{fig:canonization}. 
The relative positioning of the \gls{bn} and linear layers, as well as the ReLU nonlinearity, determines which linear layer to update with the \gls{bn} parameters.
We evaluate attributions for pre-trained VGG-16 \cite{Simonyan2015Very} and ResNet18 \cite{He2016Deep} models in our experiments (Section~\ref{sec:results}), 
which both contain segments of triples of  \emph{Convolution} $\rightarrow$ \emph{\gls{bn}} $\rightarrow$ \emph{ReLU}.

Let a convolutional layer
$y_\text{conv} = w_\text{conv} \cdot x + b_\text{conv}$,
with weight and bias parameters $w_\text{conv}$, $b_\text{conv}$,
be followed by a \gls{bn} layer
$y_\text{bn} = w_\text{bn} \cdot \frac{y_\text{conv} - \mu_\text{bn}}{s_\text{bn}} - b_\text{bn}$, with
$s_\text{bn} = \sqrt{\sigma^2_\text{bn} + \varepsilon}$. 
Here, $\mu_\text{bn}$ is the running mean and $\sigma^2_\text{bn}$ the running variance\footnote{$\mu_\text{bn}$ and $\sigma^2_\text{bn}$ parameters are fixed after training~\cite{Ioffe2015Batch}.},
$w_\text{bn}$, $b_\text{bn}$ are learnable parameters, and $\varepsilon$ a small stabilizing constant.
Following the derivations in \cite{Binder2020Notes},
the parameters $w_\text{c}$ and $b_\text{c}$ of the merged layer $y_\text{bn} = w_\text{c} \cdot x + b_\text{c}$ are then obtained as 
\begin{equation}
\begin{split}
w_\text{c} = \frac{w_\text{bn}}{s_\text{bn}} \cdot w_\text{conv} ~; \qquad
b_\text{c} = \frac{w_\text{bn}}{s_\text{bn}} \cdot (b_\text{conv} - \mu_\text{bn}) + b_\text{bn}~ .
\end{split}
\label{eq:bn_canonization}
\end{equation}

We restrict our considerations to the above case,
as we employ VGG- and ResNet-type architectures in our experiments.
Note, however, that other architectures may require different approaches for canonization~\cite{Binder2020Notes}.

\subsection{Quantifying the Quality of Explanations}
\label{sec:methods:quantifying}
We aim to evaluate the effects of model canonization on explanations not only qualitatively but also quantitatively. Therefore, for our experiments, we selected two measures that quantify two different major properties of explanations: 
\emph{Attribution Localization} \cite{Kohlbrenner2020Towards} measures the inside-total ratio $\mu$ of the sum of positive attributions inside the ground truth bounding box(es) ($R_\text{in}^+$) vs. the sum of all positive attributions ($R_\text{total}^+$) \wrt~the target label:
\begin{equation}
    \mu = \frac{R_\text{in}^+}{R_\text{total}^+}
\label{eq:localization}
\end{equation}

This measure assumes that the model prediction is based on the object within the bounding box (which can safely be assumed in most cases~\cite{Kohlbrenner2020Towards}), and it is maximized if all positive attribution quantity for the target class is concentrated within that area.
We complement this localization test with
\emph{Input Perturbation Testing} \cite{Bach2015Pixel, Samek2017Evaluating},
which does not require additional ground truth annotations.
This measure creates a ranking of input regions by sorting per-region attributions in descending order,
and successively perturbs those regions in that order.
After each perturbation, the model is evaluated on the perturbed data sample. The effect of the perturbation on the model \wrt~the original sample is recorded as
\begin{align}
    x^{(k)} &= g(x^{(k-1)}, r)\nonumber\\
    \text{IP}(x, k) &= f(x^{(0)}) - f(x^{(k)})~,
\end{align}
where $g$ is a function applying the $k$th perturbation step on $x^{(k-1)}$ according to the attribution-ranked regions $r$ that correspond to the input sample $x$.
Then, $x^{(k)}$ is the sample after $k$ perturbation steps and $x^{(0)} = x$.
This measure is maximized if the regions ranked first in $r$ according to the attribution map lead to the sharpest decrease in the output probability of the target class,
and therefore measures the faithfulness of the explaining attribution map to the prediction function.

In our experiments, we compute and evaluate attributions for a sample's true class by perturbing single-pixel regions.
We replace perturbed pixels with values drawn from a uniform distribution of all possible pixel values.

\begin{figure*}[h]
    \centering
    \includegraphics[width=.9\textwidth]{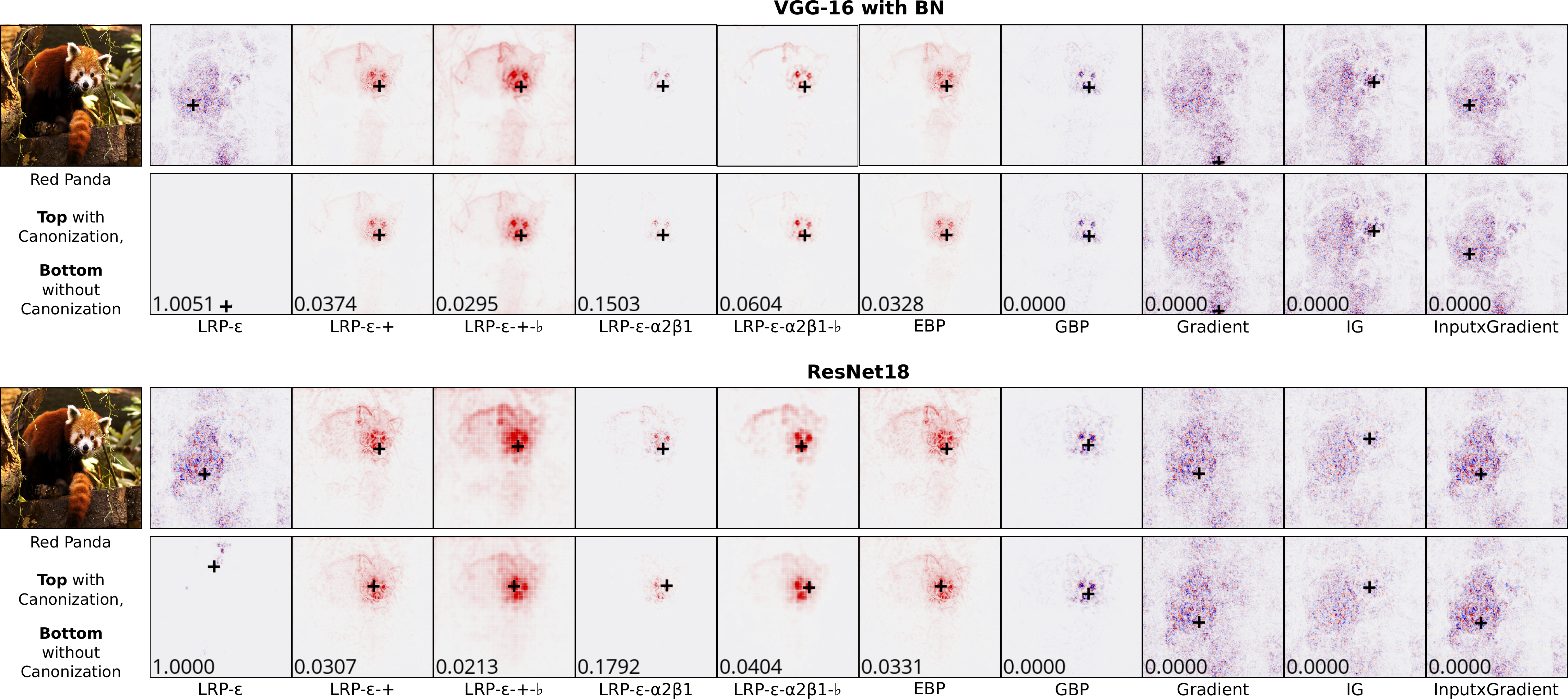}
    \caption{Explanations for VGG-16 (\emph{top}) and ResNet18 (\emph{bottom}) with and without network canonization in comparison for multiple attribution methods.
    Black crosses localize the highest attributed value per map.
    Red pixels receive positive attribution values,
    blue pixels receive negative scores.
    The cosine distances between both explanations
    from the original and the canonized model are noted in the bottom left corner of each attribution map without model canonization. Other than gradient-based approaches, rule-based attribution approaches are affected by structural changes resulting in an otherwise equivalent model.}
    \label{fig:explanations_example}
\end{figure*}

\section{Results and Discussion}
\label{sec:results}

In this section, the qualitative effects of canonization are evaluated with a visual inspection of the explanations for single samples, while the quantitative effects are measured via the two
methods introduced in Section~\ref{sec:methods:quantifying}.
For our results, two commonly used model architectures are tested. We evaluate a VGG-16 model with \gls{bn} after each convolutional layer and a ResNet18 model, both with pre-trained weights from the Pytorch%
model zoo, on a subset of samples derived from 20 randomly picked ILSVRC2012 \cite{russakovsky2015imagenet} classes\footnote{ 
 'airship',
 'balloon',
 'banana',
 'barometer', 
 'binder',
 'bison',
 'broccoli',
 'electric guitar',
 'electric switch',
 'freight car',
 'go-kart',
 'hourglass',
 'isopod',
 'ladybug', 
 'red panda',
 'reel',
 'Shih-Tzu',
 'tiger',
 'toucan',
 'volleyball'
}, counting 9758 samples with ground truth bounding boxes in total.
We compute attributions for
variants of \gls{lrp}~\cite{Bach2015Pixel}
(\cf \cite{Kohlbrenner2020Towards} for details),
\gls{ebp}~\cite{zhang2018top},
\gls{gbp}~\cite{Springenberg2015GBP},
\gls{itg}~\cite{Shrikumar2016Not},
\gls{ig}~\cite{sundararajan2017axiomatic}
and the gradient itself
using \texttt{zennit}~\cite{Anders2021Software} and its \texttt{Canonizer} implementations for comparison.

An example of the explanations for the canonized and non-canonized models can be seen in Figure \ref{fig:explanations_example}, for VGG-16 \emph{(Top)} and ResNet18 \emph{(Bottom)}.
As expected, attributions of Gradient, \gls{ig}, and \gls{itg} do not visually change with canonization, which is further confirmed by the cosine-difference between attributions, since neither gradients nor inputs are altered by canonization (although slight numerical differences may arise in practice).
For the modified backpropagation methods, however, differing results can be observed: \gls{gbp} shows no difference between canonized and non-canonized models, as it only alters the backpropagation through ReLU activations \cite{Springenberg2015GBP}. But the number of ReLU activations and their placement is not affected by the canonization applied here, so that the equal results are expected. However, \gls{ebp} and \gls{lrp}-based methods show significant differences between canonized and non-canonized models:
For \gls{lrp}-$\varepsilon$, a clear advantage can be detected for both canonized models, as the explanations for the non-canonized models do not reflect any recognizable input features. Note that this result strongly differs from our observations for \gls{itg}, so that claims of \gls{itg} and LRP-$\varepsilon$ being equal \cite{Shrikumar2016Not} do not hold here. 
But for the other \gls{lrp} configurations, differences are not as obvious (although still visible), and explanations of canonized and non-canonized models show a broad similarity. It seems that, by only relying on a visual inspection, fine-grained comparisons between similar \gls{xai} methods cannot be made in general, and the perceived differences can thus not be linked to a statement about a disparity in quality. Therefore, independent and quantitative measures are essential for assessing the effects of canonization.

\begin{figure}[h]
    \centering
    \includegraphics[width=.9\columnwidth]{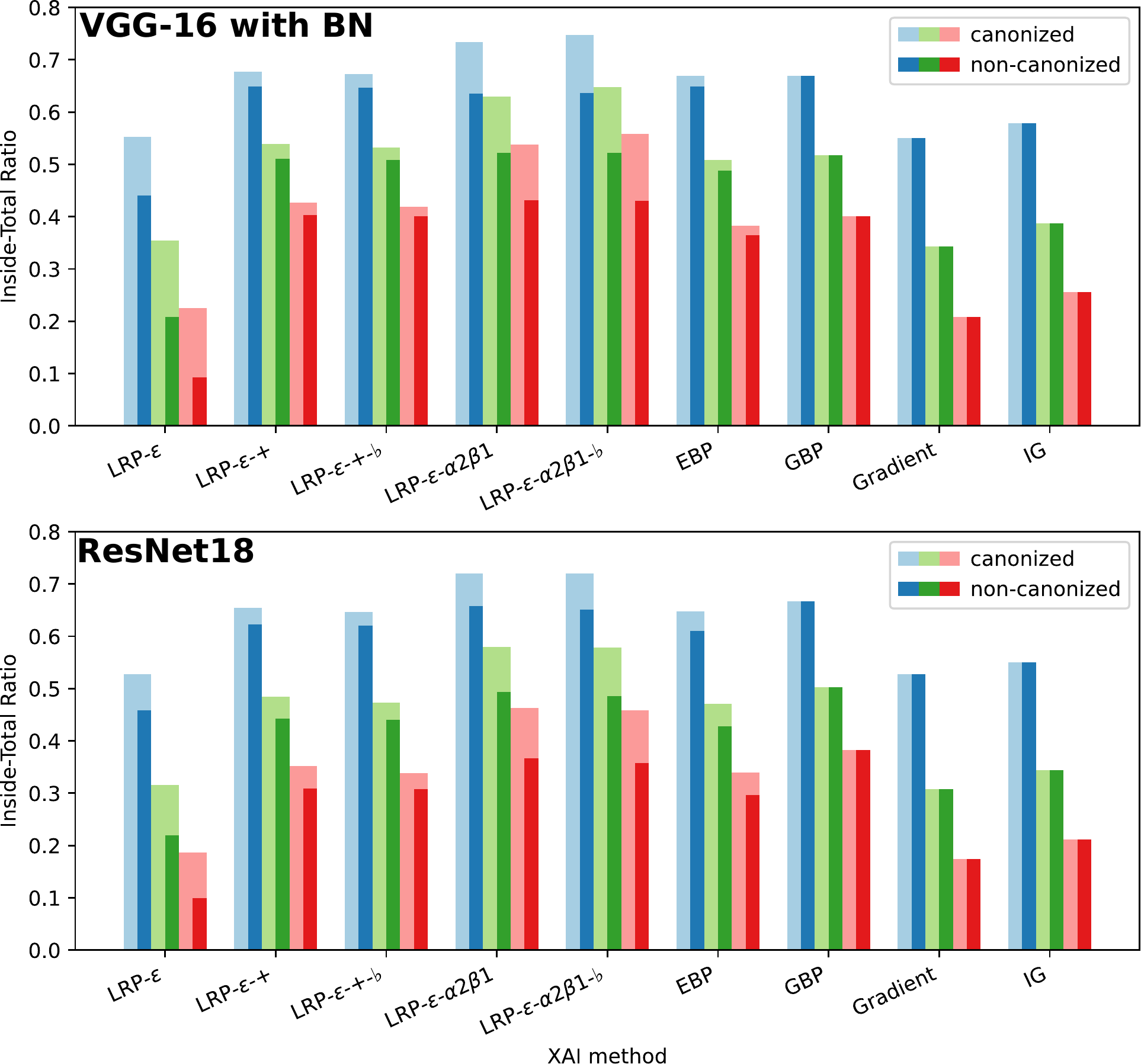}
    \caption{Comparison of the attribution localization scores for the canonized and the non-canonized VGG-16 with \gls{bn} (\emph{top}) and ResNet18 (\emph{bottom}). Bars show the mean localization scores over all labelled objects from the analyzed ILSVRC2012 subset.
    Blue bars show the score for \emph{all} objects, while green and red bars show scores for objects with bounding boxes smaller than 50\% and 25\% of the image size respectively. Different values are evaluated to ensure result independence from bounding box size, see \cite{Kohlbrenner2020Towards}.
    Color intensity indicates canonization status.}
    \label{fig:combined_localization}
\end{figure}

For this purpose, we employ two quantitative metrics focusing on different aspects of explanation quality. 
In the localization tests (see Figure \ref{fig:combined_localization}), 
explanations of \gls{lrp} methods and \glsdesc{ebp} achieve a higher inside-total ratio $\mu$ for the canonized models (\emph{Top}: VGG-16, \emph{Bottom}: ResNet18),
indicating that these explanations increase focus on the expected (bounding box) area and that canonization is therefore clearly beneficial. Note that the relative improvement in localization score is especially significant for the smaller bounding boxes where generally there is a much higher potential for attributions to be off-target.
For the other explanation methods, no changes in localization scores can be observed.

\begin{figure}
    \centering
    \includegraphics[width=.9\columnwidth]{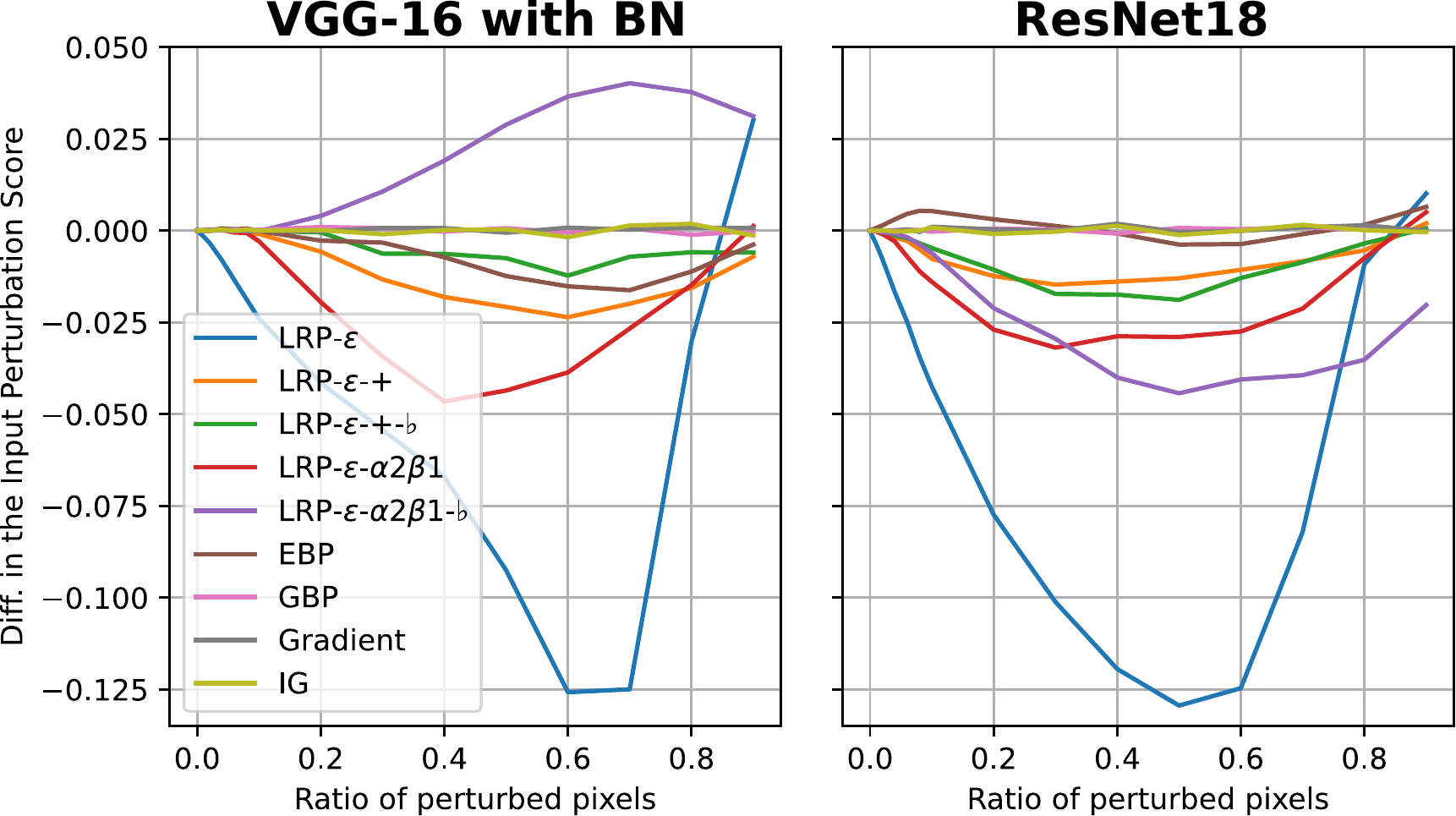}
    \caption{Differences in the input perturbation scores between the canonized and the non-canonized model.
    Scores have been obtained for the VGG-16 (\emph{left}) and the ResNet18 (\emph{right}) models on the ILSVRC2012 subset,
    using uniform sampling for replacing pixels. Negative values are associated with increased faithfulness score of the canonized model.}
    \label{fig:combined_pixelflipping}
\end{figure}

Similarly, 
the difference in input perturbation test scores between canonized and non-canonized models shown in Figure \ref{fig:combined_pixelflipping} (\emph{Left}: VGG-16, \emph{Right}: ResNet18) indicate an improvement in modified backpropagation explanation faithfulness to model decisions with canonization. 
In accordance with our qualitative findings (\cf Figure \ref{fig:explanations_example}), 
for both models, 
the input perturbation scores change the most for \gls{lrp}-$\varepsilon$.
Since \gls{lrp}-$\varepsilon$ favors sensitivity to model parameters over clarity of visual representation, canonization seems to improve explanations by correctly incorporating model parameters for explaining the model decisions.
Surprisingly, for \gls{lrp}-$\varepsilon$-$\alpha2\beta1$-$\flat$ the score for the attributions of the canonized VGG-16 is worse,
while for the ResNet18 canonization improves the attribution faithfulness.
The $\flat$-rule employed here introduces imprecision by smoothing in favor of readability. 
Thus, the effect of the $\flat$-rule is especially noticable when, as in our case, only single pixels are perturbed at a time, 
as opposed to larger regions. 
For \gls{lrp}-$\varepsilon$-$\alpha2\beta1$-$\flat$, 
the sensitivity to correct parameter incorporation is therefore decreased, 
leading to diverging results between model architectures.
Despite showing visual changes in Figure \ref{fig:explanations_example} and improved localization scores in Figure \ref{fig:combined_localization}, the effect of canonization on \gls{ebp} explanations seems ambiguous in terms of faithfulness.
In contrast, the difference for gradient-based methods, as well as for \gls{gbp}, is close to zero for both VGG-16 and ResNet18, as can be expected due to the corresponding explanations (and gradients) not being affected by canonization.
Note that --- opposed to to the localization scores and qualitative results --- the difference is not exactly zero as per the random choice of replacement values in the perturbation function inducing small variations between canonized and non-canonized models' evaluation.

\section{Conclusion}

We confirmed that canonization of \gls{bn}-layers does not affect gradient-based explanation methods, but can lead to significant improvements for modified backpropagation methods without specified rules for \gls{bn} layers. While visual differences may be apparent,
we observed that qualitative evaluations are generally not sufficient for determining whether canonization leads to better explanations. Utilizing two quantitative measures, testing the \emph{localization} and \emph{faithfulness} of explanations, respectively, we found that canonization significantly improves both properties in general --- given that the explanation method is affected by canonization at all. In doing so, canonization can provide attributions that reliably represent a model's behavior, playing an important role in ensuring model trustworthiness via explanations.

\bibliographystyle{IEEEbib}
\bibliography{main_20220214115603}
\end{document}